\documentclass[conference]{IEEEtran}
\IEEEoverridecommandlockouts
\usepackage{cite}
\usepackage{amsmath,amssymb,amsfonts}
\usepackage{graphicx}
\usepackage{textcomp}
\usepackage{algorithm}
\usepackage{algpseudocode}
\usepackage{xcolor}
\usepackage{bm}
\usepackage{natbib}
\usepackage{url}
\def\BibTeX{{\rm B\kern-.05em{\sc i\kern-.025em b}\kern-.08em
    T\kern-.1667em\lower.7ex\hbox{E}\kern-.125emX}}
\begin{document}

\title{Efficient Differentially Private Fine-Tuning of Diffusion Models\\
\thanks{$^*$This work was performed while A.L. was an intern at MERL.}
%supported by Mitsubishi Electric Research Laboratories (MERL), 201 Broadway, Cambridge, MA 02139, USA.}
}

% \author{
%     \IEEEauthorblockN{Keshav Bimbraw\IEEEauthorrefmark{1}\IEEEauthorrefmark{4}, Jing Liu\IEEEauthorrefmark{1},
%     Ye Wang\IEEEauthorrefmark{1}, Toshiaki-Koike Akino\IEEEauthorrefmark{1}, and
%     Kieran Parsons\IEEEauthorrefmark{1}}
%     \IEEEauthorblockA{\IEEEauthorrefmark{1}Connectivity and Information Processing, Mitsubishi Electric Research Laboratories, Cambridge, MA, USA\\
%     \{bimbraw, jiliu, yewang, koike, parsons\}@merl.com}
%     \IEEEauthorblockA{\IEEEauthorrefmark{4}Robotics Engineering, Worcester Polytechnic Institute, Worcester, MA, USA\\
%     \{kbimbraw\}@wpi.edu}
% }
%\footnote{Work done during internship at MERL}
\author{
    \IEEEauthorblockN{Jing Liu\IEEEauthorrefmark{2}, Andrew Lowy\IEEEauthorrefmark{1}\IEEEauthorrefmark{2}, Toshiaki Koike-Akino\IEEEauthorrefmark{2}, Kieran Parsons\IEEEauthorrefmark{2}, Ye Wang\IEEEauthorrefmark{2}}
    \IEEEauthorblockA{
     \IEEEauthorrefmark{2}Mitsubishi Electric Research Laboratories (MERL),
Cambridge, MA 02139, USA\\
\IEEEauthorrefmark{1}Wisconsin Institute for Discovery, University of Wisconsin-Madison, Madison, WI 53715, USA\\
     \{jiliu, koike, parsons, yewang\}@merl.com, alowy@wisc.edu}
}

\maketitle

\begin{abstract}
The recent developments of Diffusion Models (DMs) enable generation of astonishingly high-quality synthetic samples. Recent work showed that the synthetic samples generated by the diffusion model, which is pre-trained on public data and fully fine-tuned with differential privacy on private data, can train a downstream classifier, while achieving a good privacy-utility tradeoff. However, fully fine-tuning such large diffusion models with DP-SGD can be very resource-demanding in terms of memory usage and computation. In this work, we investigate Parameter-Efficient Fine-Tuning (PEFT) of diffusion models using Low-Dimensional Adaptation (LoDA) with Differential Privacy. We evaluate the proposed method with the MNIST and CIFAR-10 datasets and demonstrate that such efficient fine-tuning can also generate useful synthetic samples for training downstream classifiers, with guaranteed privacy protection of fine-tuning data. Our source code will be made available on GitHub. 
\end{abstract}


\section{Introduction}

Differentially Private (DP) training, via DP-SGD \citep{DP_SGD}, of a task-specific classifier is commonly used for protecting the privacy of sensitive training data. However, it often comes with a significant cost in terms of model utility. The recent developments of Diffusion Models (DMs) enable generation of astonishingly high-quality synthetic samples, which can be used to train downstream classifiers. However, the synthetic samples generated by diffusion models do not inherently preserve training data privacy. In fact, \citet{291199} demonstrates that diffusion models memorize individual images from their training data and emit them at generation time. It is vital to train the diffusion model with privacy guarantees. The recent work of \citet{ghalebikesabi2023differentially} showed that the synthetic samples generated by the diffusion model, which is fully fine-tuned with differential privacy, can train a downstream classifier that achieves very good utility. However, fully fine-tuning such large diffusion model with DP-SGD can be very resource-demanding. 

Parameter-Efficient Fine-Tuning (PEFT) updates only a small set of model parameters, which may be a subset of the existing model parameters or a set of newly added parameters, can greatly reduce the computation and memory costs, and has become popular in fine-tuning Large Language Models (LLMs). The most widely-used PEFT approach is Low-Rank Adaptation (LoRA), which constrains the updates of weights to be low-rank. In LLMs, the various modules are mostly built from linear layers, where it is straightforward to apply LoRA to the weight matrix. However, in Diffusion Models, there are many convolutional layers instead of linear layer, and the appropriate generalization of ``Low-Rank Adaptation'' becomes somewhat unclear when applied to convolutional layers. For instance, in the official implementation of LoRA\footnote{\url{https://github.com/microsoft/LoRA}}, the weight parameters of the 2D convolutional layer with dimension $(\text{out\_channels, in\_channels, kernel\_size, kernel\_size})$ is reshaped/viewed as a 2D matrix with shape $(\text{out\_channels} \times \text{kernel\_size}, \text{in\_channels} \times \text{kernel\_size})$, and the low-rank adapter is applied to this giant matrix. The recent work \citep{liu2023loda} proposed Low-Dimensional Adaptation (LoDA), which is a generalization of LoRA from a linear low-rank mapping to a nonlinear low-dimensional mapping, which is more suitable for convolutional layers than LoRA. We will discuss the details of LoDA applied to convolutional layers in Section~\ref{section:unet}.

\begin{figure*}[!h]
    \centering
    \includegraphics[width=1\textwidth]{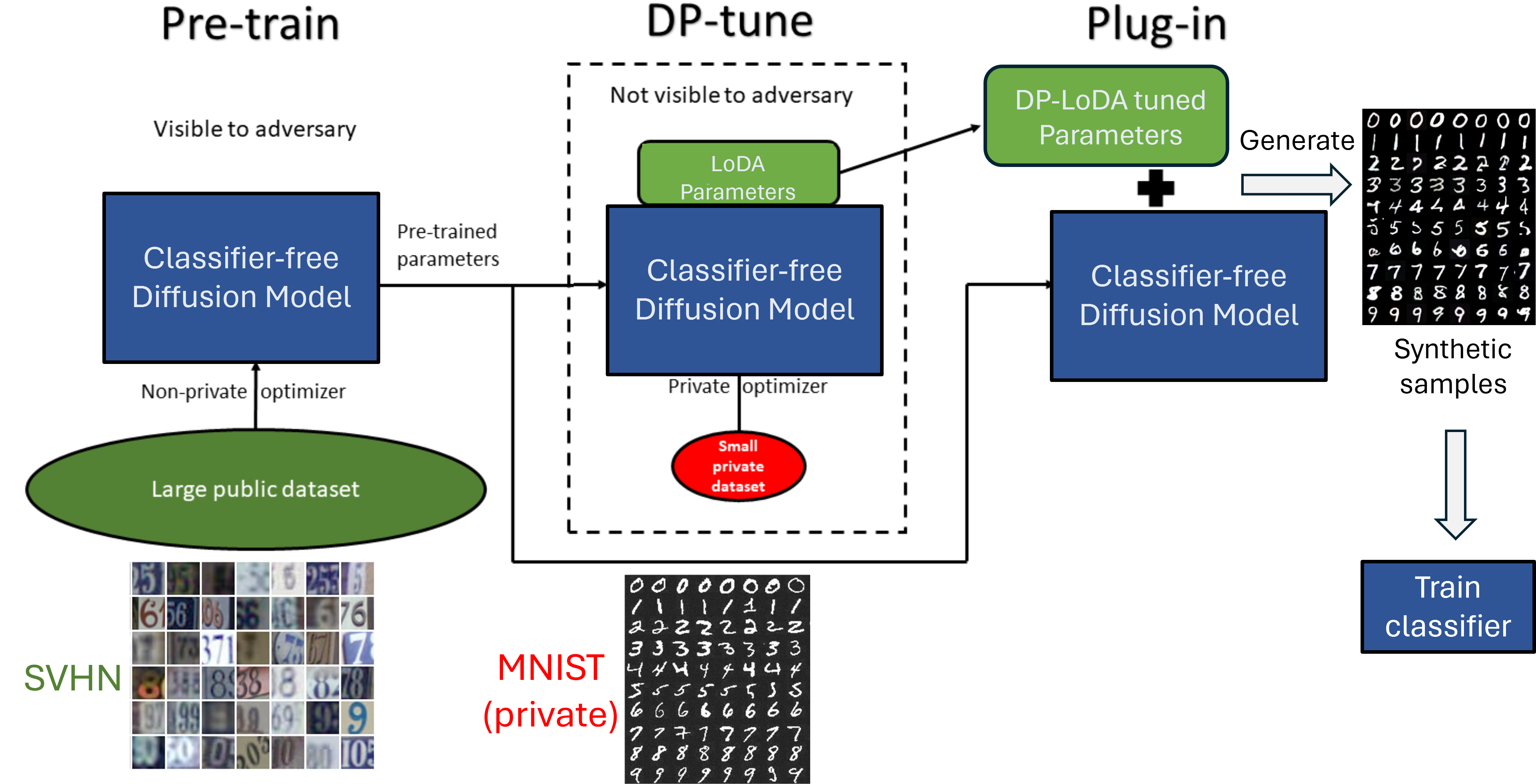}
    \vspace{2pt}
    \caption{An illustration of the proposed DP-LoDA framework for training a downstream classifier with differential privacy. Figure is modified from \citet{yu2022differentially}. }
    \label{fig:schematic}
   % \vspace{6pt}
\end{figure*} 

This motivates us to investigate whether using LoDA to fine-tune Diffusion Models with Differential Privacy (a.k.a. DP-LoDA) can also generate synthetic samples for training a downstream classifier with good utility. We notice an interesting concurrent work \citep{lyu2023differentially} that uses a Latent Diffusion Model to reduce the number of fine-tuning parameters, and they further propose to fine-tune the attention module only\footnote{In the latest version of \citet{lyu2023differentially}, we notice the authors tried LoRA fine-tuning, but the FID score of generated synthetic samples is worse than fully fine-tuning the attention module, and the downstream classification accuracy is not reported for LoRA, and remains unclear.}. It is an interesting future direction to investigate DP-LoDA combined with Latent Diffusion Models. Note that the high-level idea of combining DP with PEFT was proposed in \citet{yu2022differentially} for LLMs, but with the purpose of only fine-tuning the LLM. No synthetic samples were generated for training a downstream classifier in that work.

It is also worth mentioning that there is a large body of work using conventional generative models, such as Generative Adversarial Networks (GANs) \citep{GAN} or Variational AutoEncoders (VAEs) \citep{Kingma_2019} to generate differentially private synthetic samples \citep{DPGAN, DP_CGAN, PATE_GAN, gs-wgan, dpmerf, dphp, dp_sinkhorn, Liew0U22, DP-MEPF, harder2021dp, acs2018differentially, jiang2022dp, pfitzner2022dpd}. We compare with two recent approaches (DP-MEPF and DP-MERF) in our experiments.

In the following sections, we first introduce the overall framework of DP-LoDA, and the details of LoDA for convolutional layers. We then conduct experiments on benchmark datasets and compare with state-of-the-art methods. Finally, we conclude and discuss future directions as well as potential broader impact.

\begin{figure*}[!h]
    \centering
    \includegraphics[width=1\textwidth,trim={20 60 30 0},clip]{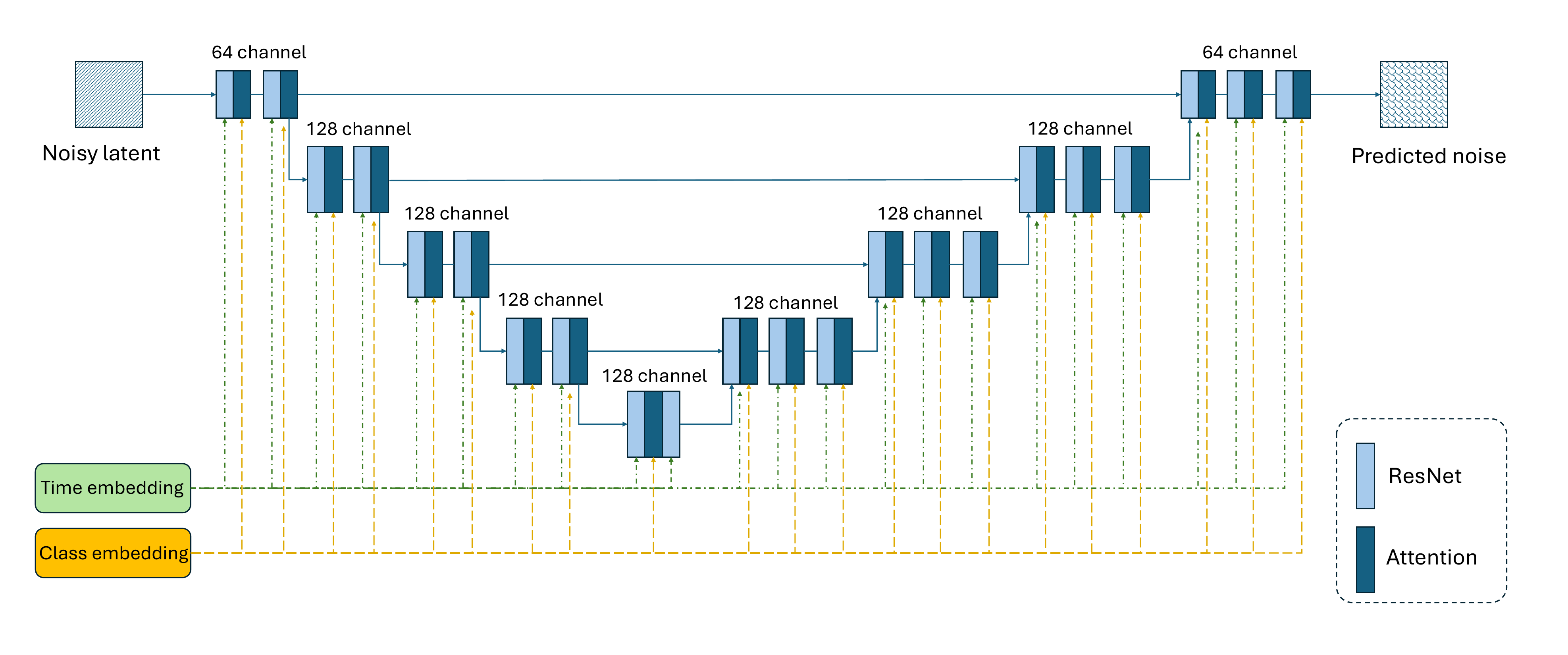}
    \caption{Classifier-free diffusion model structure. }
    \label{fig:unet}
\end{figure*}

\section{Efficiently fine-tuning of diffusion models via DP-LoDA}\label{sec:method}

The overall framework of the proposed \textbf{DP-LoDA} method is illustrated in Fig. \ref{fig:schematic}. First, the classifier-free diffusion model \citep{ho2021classifier} is pre-trained on some public dataset. Next, LoDA adapters are attached to the Diffusion Model and are differential privately fine-tuned (via DP-SGD) on a smaller private dataset. The original parameters of the pre-trained diffusion models are frozen during this process. Then, the DP-LoDA fine-tuned, classifier-free diffusion model is used to generate synthetic samples with class conditioning. Finally, the downstream classifier is trained with the generated synthetic samples.

\begin{figure}[h]%{0.3\textwidth}
    \centering
    %\vspace{-5mm}
    \includegraphics[width=0.45\textwidth,trim={0 20 0 0},clip]{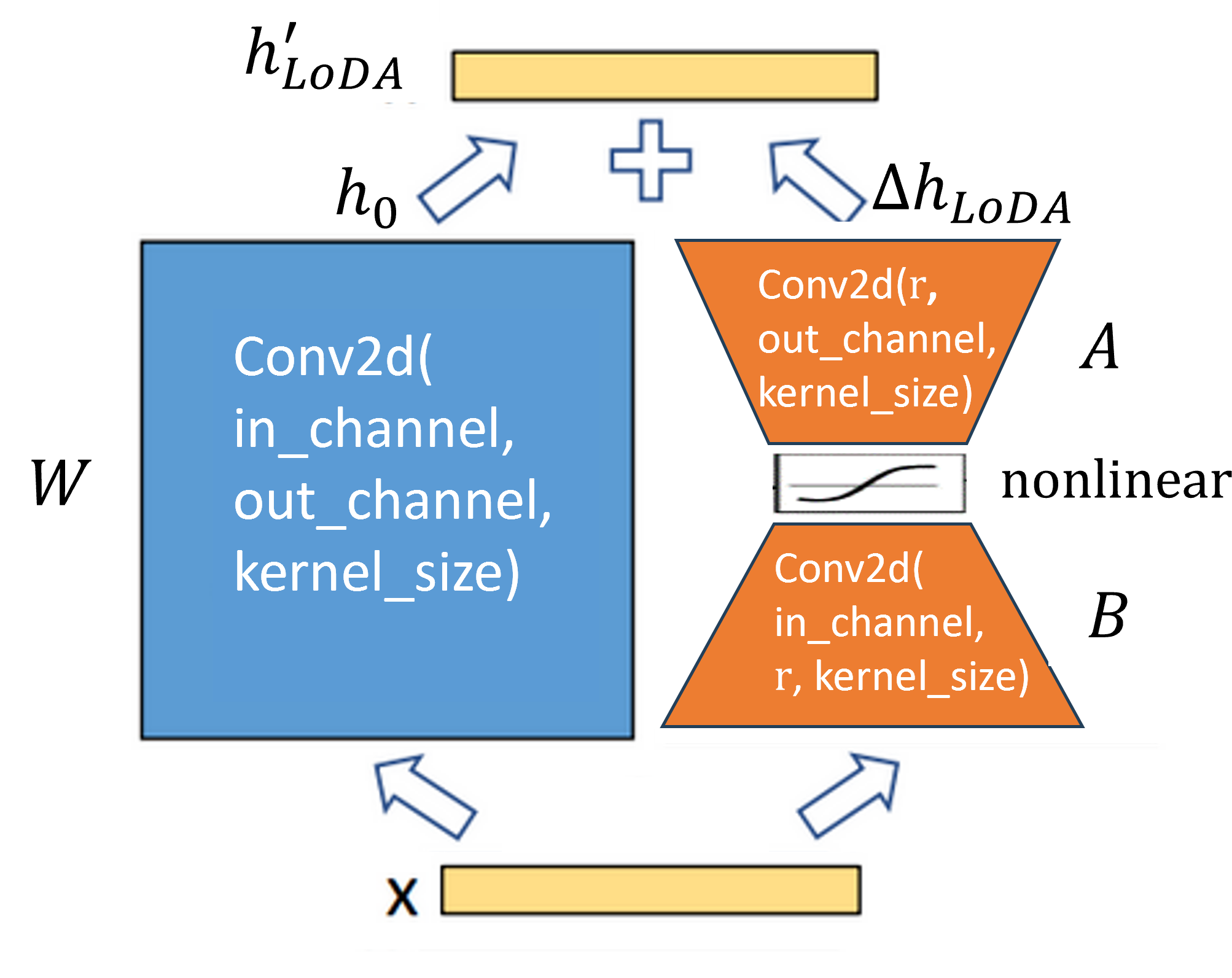}
    \caption{Low-Dimensional Adaptation (LoDA) for convolutional layer.}
    \label{fig:loda}
\end{figure}

\subsection{Illustration of diffusion model and LoDA adapter}
\label{section:unet}

The classifier-free diffusion model~\citep{ho2021classifier} that we used for image generation is adapted from a public codebase\footnote{\url{https://github.com/coderpiaobozhe/classifier-free-diffusion-guidance-Pytorch
}}. 
Its U-Net structure is illustrated in Fig. \ref{fig:unet}, which contains 21 attention modules and 22 ResNet modules, where most of the layers (including the query/value/key projection layers) are convolutional layers. 
The total number of parameters in the U-Net base model is $10.42$M. Following the idea of Low-Dimensional Adaptation (LoDA), we attach a low-dimensional convolutional adapter in parallel with the original convolutional layer. As illustrated in Fig.~\ref{fig:loda}, it consists of 1) a convolutional layer $A$ which has output channel size $r$ that is much smaller than the input channel size; 2) one or more nonlinear function(s); 3) a convolutional layer $B$ which maps the $r$ channel latent back to the original output channel size. 
During the LoDA fine-tuning, the original convolutional layer $W$ is frozen and only the LoDA adapter is fine-tuned. As the value of $r$ is much smaller than the original input/output channel size, the number of tunable parameters in $A$ and $B$ is much smaller than the original convolutional layer $W$.

\section{Experiments}\label{sec:experiment}

We first compare the proposed method with the widely used standard DP method, which simply trains the classifier using DP-SGD. We also compare with several state-of-the-art generative model based approaches, such as DP-Diffusion \citep{ghalebikesabi2023differentially} that fully fine-tunes the Diffusion Model, DP-LDM \citep{lyu2023differentially} that fine-tunes the attention module of the Latent Diffusion Model, DP-MEPF \citep{DP-MEPF}, and DP-MERF \citep{dpmerf}. As a reference, we also report the accuracy of the classifier when it is trained on private data without any privacy protection, denoted as ``No DP''.

We consider MNIST \citep{lecun2010mnist} and CIFAR-10 \citep{krizhevsky2009learning} as the private datasets, and leverage SVHN~\citep{netzer2011reading} and ImageNet32 \citep{deng2009imagenet} as the corresponding public datasets. We also consider the setting that each class of CIFAR-10 has only 1\% of the original training samples, to simulate the situation of a small private dataset in practice, which may often be the case for private medical or biosignal datasets, for example.

For our proposed DP-LoDA, the dimension $r$ is set to 4 in all of the experiments. \textcolor{black}{We use LeakyReLU with negative slope of $0.1$ as the nonlinear function between $A$ and $B$ in LoDA.} Our implementations of the LoDA adapter for the convolutional layer is modified from a public codebase\footnote{\url{https://github.com/cloneofsimo/lora}}, and the DP accounting tool we used is from \citet{bu2023differentially}. 
We fine-tune for 200 epochs on the CIFAR-10 training set, and 100 epochs on the MNIST training set. We report the downstream classification accuracies on the the true CIFAR-10/MNIST testing set, under various privacy levels.

\subsection{Experiments on CIFAR-10}

In this first set of experiments, we use the training set of CIFAR-10 \citep{krizhevsky2009learning} as private dataset, and leverage ImageNet32 \citep{deng2009imagenet} as the corresponding public dataset. However, ImageNet32 has 1000 classes, and the 10 classes of CIFAR-10 are not exactly covered by the ImageNet32 classes. Motivated by \citet{DBLP:conf/iclr/HuangME0021}, which identifies the ImageNet classes that are close to CIFAR-10 classes, we identified 10 classes of ImageNet32 that are similar to CIFAR-10 classes, listed in Table~\ref{tab:class_mapping}.
\begin{table}[h]
    \centering
        \caption{The 10 ImageNet classes similar to the CIFAR-10 classes.}
       % \vspace{5pt}
    \begin{tabular}{cc}
    \hline
        CIFAR-10 Class & ImageNet32 Class \\
        \hline
        Airplane & Airliner\\
         Car & Wagon\\
         Bird & Humming Bird\\
         Cat & Siamese Cat \\
         Deer & Hartebeest \\
         Dog & Golden Retriever\\
         Frog & Tailed Frog\\
         Horse & Sorrel Horse\\
         Ship & Container Ship \\
         Truck & Trailer Truck \\
         \hline
    \end{tabular}

    \label{tab:class_mapping}
\end{table}

However, for those corresponding 10 ImageNet32 classes, there are only 13k training samples in total, and we empirically found that they are not sufficient to train the Classifier-free Diffusion Model. Since ImageNet32 has more than 1281k training samples overall,
we propose the following pre-training procedure to leverage the entire ImageNet32 dataset:

\textbf{Step 1:} Pre-train the diffusion model without class conditioning using all ImageNet32 samples; 

\textbf{Step 2:} Continue pre-training using the corresponding 10 ImageNet32 classes only, with class conditioning. 

Fig. \ref{fig:images_DM1} in the Appendix shows the images generated by the Diffusion
Model following Step 1, after pre-training for 300 epochs on all ImageNet32 training samples without class conditioning. Fig. \ref{fig:images_DM2} in the Appendix shows the images generated by the Diffusion Model following Step 2, after further pre-training of 1200 epochs with class conditioning on the corresponding 10 ImageNet32 classes. We can see that Step 2 is very effective at training class conditioning.

\subsubsection{CIFAR-10 as the Private Data}
The full training set of CIFAR-10 has 50K samples. DP-LDM generates 50K synthetic samples for training the classifier, while DP-Diffusion generates 1 million synthetic samples. In our proposed method, we follow the conventions of DP-Diffusion by generating 1 million synthetic samples. Table~\ref{tab:cifarall} shows the test accuracies (on the true CIFAR-10 testing set) of ResNet9 by each DP training method (with access to the full CIFAR-10 training set). Various levels of $(\epsilon,\delta)$-DP are reported, with $\delta$ fixed at $10^{-5}$.

First of all, we can see that DP-LDM, DP-Diffusion, and the proposed DP-LoDA significantly outperform DP-SGD, DP-MERF and DP-MEPF. When $\epsilon$ is relatively small, the proposed DP-LoDA outperforms DP-LDM. There is still some gap between parameter-efficiently fine-tuned Diffusion Model and fully fine-tuned Diffusion Model. Note that the classifier accuracy can be further improved by using the Wide ResNet (WRN-40-4) architecture instead of ResNet9, as demonstrated in~\citet{lyu2023differentially}. Further, \citet{deepmind_DPDM} showed that the accuracy can be further improved by using the augmentation multiplicity technique~\citep{de2022unlocking}. Finally, as a reference, Fig. \ref{fig:images_DM_cifarall} in the Appendix shows the images generated by Diffusion Model after DP-LoDA fine-tuning with $(\epsilon=10, \delta = 10^{-5})$ on the full CIFAR-10 training set.
\begin{table}[h]
\centering
\caption{Test accuracies (on the true CIFAR-10 testing set) of ResNet9 by each DP training method (with access to the full CIFAR-10 training set), under different levels of privacy cost $(\epsilon,\delta)$, with $\delta$ fixed to be $10^{-5}$. The values of DP-LDM, DP-MEPF and DP-MERF are from \citet{lyu2023differentially}.
}
%\vspace{5pt}
\scalebox{0.92}{
\begin{tabular}{l | c|c|c }
% &  \multicolumn{3}{c }{\textit{ResNet 9} } \\
\hline
  Method &  $\epsilon=1$ & $\epsilon=5$   & $\epsilon=10$ \\
\hline
 \text{DP-LDM} &  \text{51.3 $\pm$ 0.1}  & \text{59.1 $\pm$ 0.2} & \text{65.3 $\pm$ 0.3}\\
  \textcolor{black}{\text{\textbf{DP-LoDA}}} & \text{60.2 $\pm$ 0.2} & \text{62.2 $\pm$ 0.4} &  \text{63.5 $\pm$ 1.8} \\
   \textcolor{black}{\text{DP-Diffusion}} & \text{66.3 $\pm$ 0.4}  & \text{69.6 $\pm$ 0.2} &  \text{69.7 $\pm$ 1.4}\\
   \textcolor{black}{\text{DP-SGD}} & \text{36.5 $\pm$ 0.9} & \text{47.4 $\pm$ 0.9} &  \text{48.3 $\pm$ 0.2} \\  
{DP-MEPF ($\phi_1, \phi_2$)} & 28.9  & 47.9   &   48.9  \\
{DP-MEPF ($\phi_1$)}          & 29.4  & 48.5    &  51.0   \\
DP-MERF      & 13.8  & 13.4     & 13.2  \\
\hline
{No DP } & \multicolumn{3}{c}{90.7}  \\
   \hline
\end{tabular}\label{tab:cifarall}}
\end{table}

\subsubsection{1\% of CIFAR-10 as the Private Data}
In some cases of private datasets, such as medical or biosignal data, the number of training samples may be quite limited. To simulate this scenario, we conduct an experiment with only 1\% of the CIFAR-10 training set as the private dataset.
 
Table~\ref{tab:cifar1} shows the test accuracies (on the true CIFAR-10 testing set) of ResNet9 using each DP training method (with access to $1\%$ of the CIFAR-10 training set), under the common DP settings $\epsilon=1, 10$ and $\delta=10^{-5}$. First of all, we can see that DP-SGD works poorly under this setting. The gap between the accuracies of DP-Diffusion and DP-LoDA becomes very small, which aligns with the findings in the LLM literature that PEFT approaches perform similarly as full fine-tuning when the fine-tuning dataset is small. More interestingly, \textit{even at a very small privacy cost of $(\epsilon =1,\delta =10^{-5})$, both DP-Diffusion and DP-LoDA outperform the accuracy of the classifier trained without DP} (``No DP''). Finally, as a reference, Fig. \ref{fig:images_DM_cifar1} in the Appendix shows the generated images by Diffusion Model after {DP-LoDA} fine-tuning with $(\epsilon=10, \delta = 10^{-5})$ using 1\% of the CIFAR-10 training set. Even with such a very limited private dataset, the quality of the generated synthetic samples are still good.

\begin{table}[t]
\centering
\caption{Test accuracies (on the true CIFAR-10 testing set) of ResNet9 by each DP training method (with access to 1\% CIFAR-10 training set), under different levels of privacy cost $(\epsilon,\delta)$, with $\delta$ fixed to be $10^{-5}$.
}
%\vspace{6pt}
\scalebox{1.0}{
\begin{tabular}{l | c|c }
 %\textit{ResNet 9}
 \hline
  Method&  $\epsilon=1$    & $\epsilon=10$ \\
\hline

  \textcolor{black}{\textbf{\text{DP-LoDA}}} & \text{54.2} &  \text{53.6}  \\
   \textcolor{black}{\text{DP-Diffusion}} & \text{54.6} &  \text{55.9}  \\
   \textcolor{black}{\text{DP-SGD}} & \text{11.5} &  \text{21.2}  \\  
   \hline
{No DP } & \multicolumn{2}{c}{52.5}  \\%{} &       \\

\hline
\end{tabular}
\label{tab:cifar1}}
\end{table}

\subsection{Experiments on MNIST}
In this section, we consider the 60K sample training set of MNIST \citep{lecun2010mnist} as the private dataset, and we use SVHN~\citep{netzer2011reading} as the public dataset. Note that SVHN has exactly the same classes as MNIST, \textcolor{black}{and therefore we directly pre-train the diffusion model with class conditioning. Fig. \ref{fig:images_DM_svhn} in the Appendix shows the images generated by the pre-trained Diffusion Model. Fig. \ref{fig:images_DM_mnist} in the Appendix shows the images generated by Diffusion Model after DP-LoDA fine-tuning with $(\epsilon=10, \delta = 10^{-5})$ on MNIST training set. It is clear that DP-LoDA fine-tuning can successfully transfer from SVHN to MNIST.}

For the tuned Diffusion Models, we generate 60K synthetic samples for both DP-Diffusion and DP-LoDA methods. We fix $(\epsilon = 10,\delta = 10^{-5})$ and report the test accuracy of different methods in Table~\ref{tab:mnist}.

\begin{table}[!h]
\caption{Test accuracies (on true MNIST testing set) of a CNN classifier by each DP training method (with access to the full MNIST training set). The value of DP-LDM is from \citet{lyu2023differentially}.}
%\vspace{9pt}
\centering
\scalebox{1.0}{
\begin{tabular}{ll|cc}
\hline
&  Method &   $(\epsilon=10, \delta = 10^{-5})$ \\
 \hline
& DP-LDM & 94.3  \\
& \textbf{DP-LoDA}  &  95.0 \\
& DP-Diffusion & 95.9 \\
& DP-SGD & 79.3 \\
\hline
& No DP & 99.4 \\
\hline
\end{tabular}
\label{tab:mnist}
}
\end{table}

DP-Diffusion, DP-LDM, and the proposed DP-LoDA methods significantly outperform DP-SGD, and achieve very good utility. DP-LoDA performs slightly better than DP-LDM and close to DP-Diffusion. 

Since SVHN dataset has exactly the same classes as MNIST, as a reference, we also tried directly training the classifier on the SVHN dataset and testing on the MNIST testing set, which yields a classification accuracy of only 60.22\%.

\section{Conclusion and Future Work}

We have proposed and evaluated DP-LoDA for efficiently fine tuning a diffusion model with differential privacy, which is then used to generate synthetic samples for training a downstream classifier. We have evaluated the proposed method on the MNIST and CIFAR-10 datasets, and demonstrated that such efficient fine-tuning can generate useful synthetic samples for training downstream classifiers, while protecting the privacy of the fine-tuning data. Empirical studies on the CIFAR-10 dataset further show that such parameter-efficient method gains more advantage when the private fine-tuning data is limited. An interesting future exploration is to test DP-LoDA on Latent Diffusion Models, which can further improve parameter-efficiency.

\section*{Impact Statement}
This paper presents work whose goal is to propose and investigate whether the efficient fine-tuning of vision foundation model under Differential Privacy is effective for data privacy protection and useful for training a downstream classifier. There are many positive social impacts in terms of privacy protection and computational efficiency. We do not feel there are any major negative impacts.

\bibliography{reference}
\bibliographystyle{spbasic}

%%%%%%%%%%%%%%%%%%%%%%%%%%%%%%%%%%%%%%%%%%%%%%%%%%%%%%%%%%%%%%%%%%%%%%%%%%%%%%%
%%%%%%%%%%%%%%%%%%%%%%%%%%%%%%%%%%%%%%%%%%%%%%%%%%%%%%%%%%%%%%%%%%%%%%%%%%%%%%%
% APPENDIX
%%%%%%%%%%%%%%%%%%%%%%%%%%%%%%%%%%%%%%%%%%%%%%%%%%%%%%%%%%%%%%%%%%%%%%%%%%%%%%%
%%%%%%%%%%%%%%%%%%%%%%%%%%%%%%%%%%%%%%%%%%%%%%%%%%%%%%%%%%%%%%%%%%%%%%%%%%%%%%%

%\appendix
\onecolumn
\newpage
\section{Appendix}

\begin{figure*}[h]
\centering
\begin{minipage}{.41\textwidth}
  \centering
  \includegraphics[width=.81\linewidth]{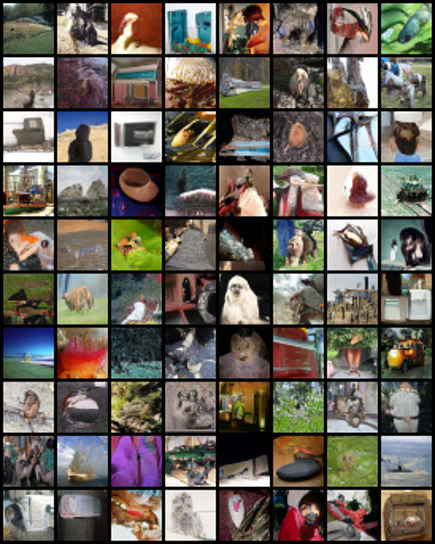}
  \caption{Generated images by Diffusion Model from Step 1 of pre-training (epoch 300).
}
  \label{fig:images_DM1}
\end{minipage}%
\hspace{1.5em}
\begin{minipage}{.41\textwidth}
  \centering
  \includegraphics[width=.81\linewidth]{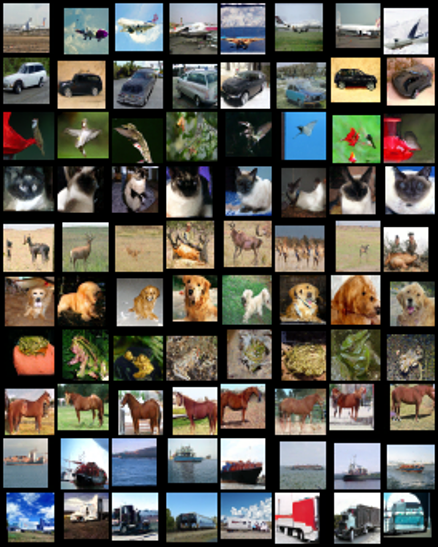}
  \caption{Generated images by Diffusion Model from Step 2 of pre-training (epoch 1200).}
  \label{fig:images_DM2}
\end{minipage}
\end{figure*}

\vspace{15pt}

\begin{figure*}[h]
\centering
\begin{minipage}{.41\textwidth}
  \centering
  \includegraphics[width=.81\linewidth]{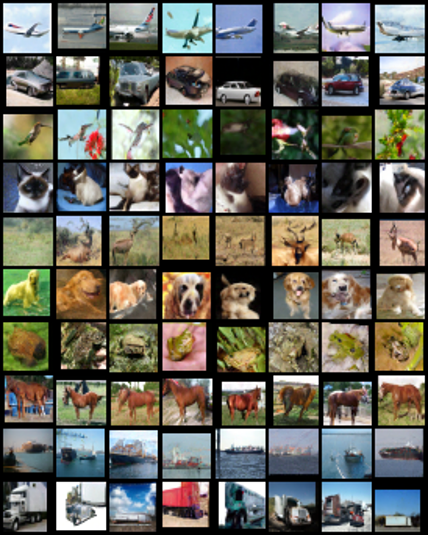}
  \caption{Generated images by Diffusion Model after DP-LoDA fine-tuning with $(\epsilon=10, \delta = 10^{-5})$ on full CIFAR-10 training set (epoch 200).
}
  \label{fig:images_DM_cifarall}
\end{minipage}%
\hspace{1.5em}
\begin{minipage}{.41\textwidth}
  \centering
  \includegraphics[width=.81\linewidth]{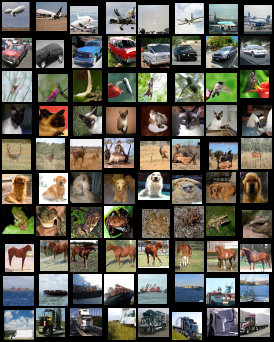}
  \caption{Generated images by Diffusion Model after DP-LoDA fine-tuning with $(\epsilon=10, \delta = 10^{-5})$ on 1\% CIFAR-10 training set (epoch 200).}
  \label{fig:images_DM_cifar1}
\end{minipage}
\end{figure*}

\begin{figure*}[h]
\centering
\begin{minipage}{.41\textwidth}
  \centering
  \includegraphics[width=.81\linewidth]{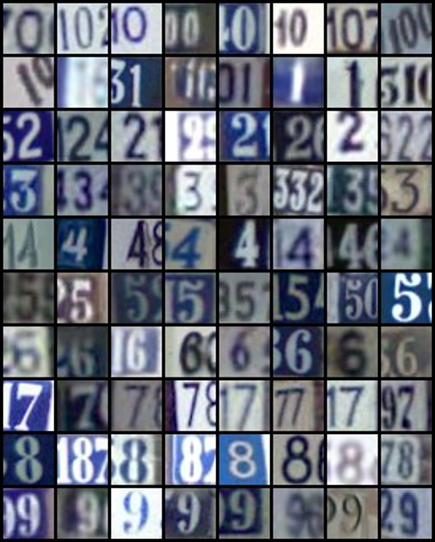}
  \caption{Generated images by Diffusion Model after pre-trained 720 epochs on SVHN dataset.
}
  \label{fig:images_DM_svhn}
\end{minipage}%
\hspace{1.5em}
\begin{minipage}{.41\textwidth}
  \centering
  \includegraphics[width=.8\linewidth]{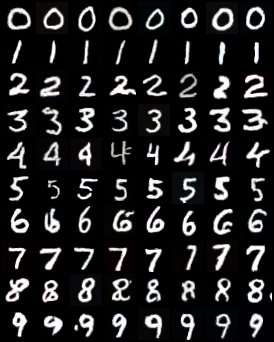}
  \caption{Generated images by Diffusion Model after DP-LoDA fine-tuning with $(\epsilon=10, \delta = 10^{-5})$ on MNIST training set (epoch 100).}
  \label{fig:images_DM_mnist}
\end{minipage}
\end{figure*}

\end{document}